\documentclass[conference]{IEEEtran}
\IEEEoverridecommandlockouts
\DeclareUnicodeCharacter{202F}{ }
\usepackage{cite}
\usepackage{amsmath,amssymb,amsfonts}
\usepackage{algorithmic}
\usepackage{graphicx}
\usepackage{textcomp}
\usepackage{booktabs}
\usepackage{tabularx}
\usepackage{xcolor}

\usepackage[numbers]{natbib}
\def\BibTeX{{\rm B\kern-.05em{\sc i\kern-.025em b}\kern-.08em
    T\kern-.1667em\lower.7ex\hbox{E}\kern-.125emX}}
\begin{document}

\title{ Trained Miniatures: Low cost, High Efficacy SLMs for Sales \& Marketing
\thanks{SuperAGI Research}
}
\makeatletter
\newcommand{\linebreakand}{%
  \end{@IEEEauthorhalign}
  \hfill\mbox{}\par
  \mbox{}\hfill\begin{@IEEEauthorhalign}
}
\makeatother
\author{
\IEEEauthorblockN{Ishaan Bhola}
\IEEEauthorblockA{
\textit{SuperAGI Research}}
\and
\IEEEauthorblockN{Harsh Nandwani}
\IEEEauthorblockA{
\textit{SuperAGI Research} }
\and
\IEEEauthorblockN{Sravanth Kurmala}
\IEEEauthorblockA{
\textit{SuperAGI Research}}
\linebreakand
\IEEEauthorblockN{Arihant Jain}
\IEEEauthorblockA{
\textit{SuperAGI Research}}
\and
\IEEEauthorblockN{Mukunda NS}
\IEEEauthorblockA{
\textit{SuperAGI Research}}

}

\maketitle

\begin{abstract}
Large language models (LLMs) excel in text generation; however, these creative elements require heavy computation and are accompanied by a steep cost. Especially for targeted applications such as sales and marketing outreach, these costs are far from feasible. This paper introduces the concept of ``Trained Miniatures" --- Small Language Models(SLMs) fine-tuned for specific, high-value applications, generating similar domain-specific responses for a fraction of the cost. \end{abstract}


\section{Introduction}

The rapid development of the Transformer architecture has been a catalyst for a paradigm shift in artificial intelligence, fundamentally revolutionizing the capabilities of Language Models (LMs). This architectural breakthrough, pioneered by Vaswani et al. (2017)\cite{vaswani2023attentionneed}, established the foundation for an unprecedented era of scale. Its success confirmed the scaling hypothesis, a governing principle which asserts a predictable, power-law connection between a model's performance and the increase in its parameter size, dataset size, and computational budget (Kaplan et al., 2020)\cite{DBLP:journals/corr/abs-2001-08361}. The unforgiving pursuit of this principle has led to the creation of Large Language Models (LLMs)—gigantic neural networks, frequently containing hundreds of billions of parameters, which exhibit emergent capabilities. Defined by Wei et al. (2022)\cite{wei2021finetuned}, these are advanced capabilities in reasoning, problem-solving, and in-context learning that do not exist in smaller models but emerge at colossal scale. These immense LLMs, whether commercialized systems or market-leading open-source initiatives, now dominate a wide range of generalized tasks, establishing a new and daunting benchmark for machine-generated content.

Though revolutionary in their potential, direct application of such large-scale LLMs is practically and strategically difficult for specialized business activities, particularly in high-volume, high-velocity business domains like sales and marketing. Even with open-source LLMs, the computational cost and inference latency of a tens- or hundreds-of-billions-parameter model become infeasible. For a business activity that wants to generate thousands or even millions of personalized communications, the capital cost and compute time render direct LLM application infeasible. More importantly, the generalist nature of such models is strategically non-ideal for highly specialized tasks where success is not in terms of Turing-like conversational ability, but in cold, hard business metrics. In such business uses, the ultimate goal is to generate near-direct effects on key performance indicators (KPIs) such as email open rates, click-through rates (CTR), positive reply rates, and, ultimately, sales pipeline creation and lead conversion rates. Asked simply by a generalist LLM, however mighty, is an untargeted and inefficacious way to generate such quantifiable results.

In this work, we introduce the idea of ``Trained Miniatures"—a technical-strategic method to build hyper-specialized, cost-effective, and privately-owned language models. The method is a two-step process that leverages the power of a large model to build a much more efficient, application-specific asset. The first step is to utilize a more powerful, larger LLM—either a state-of-the-art proprietary model consumed through API or an open-source, high-performance model—to serve as a smart data generation machine. This ``teacher" model is asked to produce a vast quantity of high-quality outputs for a specific business function. For instance, it can be instructed to generate a tailored sales email by pulling in information from a prospect's LinkedIn profile, company news, and market intelligence. This machine-generated data is then put through a strict human-in-the-loop verification process, where domain experts curate and refine the outputs, filtering for strategic relevance, quality, and precision.
This results in a ``gold standard" training dataset of thousands of flawless input-output pairs. This refined dataset is then used to fine-tune an order-of-magnitude smaller, more efficient open-source Small Language Model (SLM).

It is essential to make a clear technical distinction from traditional knowledge distillation (Hinton et al., 2015)\cite{hinton2015distillingknowledgeneuralnetwork}. Traditional distillation is a difficult process that attempts to project a ``teacher" model's whole knowledge base onto a ``student" model. This is typically done by training the student to mimic the teacher's internal output probability distributions (the ``soft targets" or logits) over the entire vocabulary. The goal is to compress the student's internal computational process into a compact representation of the teacher's. Our ``Trained Miniatures" method, in contrast, is a form of behavioral cloning by instruction fine-tuning. This is not only different; it is strategically beneficial for our stated objective: to precisely replicate the external behavior of a model for a particular domain, especially when the teacher model is unavailable to modify. This is a tremendous advantage when the teacher model is a proprietary API (where logits are not available) or a massive open-source model that one does not wish to execute in an expensive and clunky co-located training loop. Our process treats the LLM as a ``black box" data generator, completely removing the expensive and slow data generation step from the high-speed and efficient fine-tuning step. The LLM's job is not to teach its internal reasoning process, but to generate the final, polished text outputs that are exemplary demonstrations of a target ability. 

This difference emphasizes a necessary and deliberate trade-off. Our method is well-suited to cloning some particular, concrete behavior—such as writing one kind of email—but not to copying the underlying, general-purpose reasoning ability of the larger model. By restricting the SLM to training on the end-text outputs itself, it acquires the high-level patterns, structure, and vocabulary that make up a correct answer within its limited domain. It does not acquire, however, the rich, multi-step, and frequently abstract reasoning that the larger LLM may have employed to reach that answer. A Trained Miniature thus performs its specialized function with impressive dependability but without the general, flexible problem-solving capability of its much larger parent. This is not a shortcoming of the method; it is its strength. For sales and marketing applications, the final objective is the consistent, low-latency, and scalable reproduction of a high-quality output, not the production of a smaller, general-purpose reasoner. The shallow, narrow expertise is the goal.

Furthermore, these trained minatures are compared head-to-head with LLMs in an agentic fashion with the ultimate goal of managing the content generation of cold-outreach campaigns. These campaigns involve a combination of emails, LinkedIn messages and replies(if any) to these outreach attempts. The results of these Trained Miniatures add weight to the proposition of Belcak et al. \cite{belcak2025small} highlighting SLMs being the future of AI agents.

This work's value is therefore a working, reproducible, and strategically successful methodology to construct bespoke AI assets to bridge the gaping gap between raw LLM potential and the focused, ROI-driven needs of modern enterprise applications. We will demonstrate how this methodology reduces inference cost and latency by orders of magnitude, eliminates vendor lock-in, and provides organizations with full control and ownership of their AI models and underlying data. Our Trained Miniatures' performance will be tested rigorously in sales and marketing applications, with a focus on their quantitative ability to meaningfully improve the key business metrics that drive success in targeted outreach. This paper describes a realistic route to the democratization of state-of-the-art AI, enabling organizations to develop and manage powerful, specialist language models that generate measurable and substantial return on investment without the extreme cost and strategic risk of using large, general-purpose model.

\section{Literature Review}
Recent advancements in natural language processing have led to the development of Large Language Models (LLMs) and Small Language Models (SLMs). LLMs, such as Claude-4 and GPT 4o, have shown impressive abilities across many tasks. However, their high computational needs and costs create challenges for many organizations. In contrast, SLMs provide a more resource-efficient option, especially when they are fine-tuned for specific tasks. Recent research has concentrated on transferring knowledge from larger models to smaller ones by teaching the probabilistic distribution, also known as soft labels\cite{hinton2015distillingknowledgeneuralnetwork}. This section details current advancements and summarizes several research papers focused on related areas.

Bucher and Martini (2024) found that fine-tuned SLMs consistently outperform zero-shot generative AI models in text classification tasks. Their study compared models like ChatGPT (GPT-3.5/GPT-4) and Claude Opus with fine-tuned BERT-style models across various classification tasks. They discovered that task-specific fine-tuning leads to better performance \cite{bucher2024finetunedsmallllmsstill}. Similarly, Pecher et al. (2024) examined the number of labeled samples necessary for specialized small models to surpass general large models in text classification. They found that fine-tuned SLMs often need only a few samples (on average 10 to 1,000) to match or exceed the performance of larger models, depending on the dataset and task \cite{pecher2025comparingspecialisedsmallgeneral} . Later, Edwards and Camacho-Collados (2024) conducted a large-scale evaluation comparing zero- and few-shot approaches of large language models with fine-tuning smaller language models across 16 text classification datasets. Their results showed that fine-tuning smaller, more efficient language models still outperforms few-shot approaches of larger models, emphasizing the importance of fine-tuning in text classification tasks \cite{edwards2024language}. Recent work from NVIDIA \cite{belcak2025smalllanguagemodelsfuture} demonstrates that LLMs are not efficient for large-scale deployment. While benchmark scores suggest that SLMs generally perform lower than most state-of-the-art LLMs, fine-tuned SLM models show a different result. When looking at deployment needs like computational overhead, latency, and training costs, fine-tuned SLMs clearly become the better choice for large-scale deployments.

SLMs are smaller and simpler by design. This leads to higher speed and lower operating costs. They are considerably more cost-effective than LLMs in both training and deployment. Training an LLM can cost between $1 million and over $100 million, while SLMs require much less investment. For example, training a small model like Phi-2 or TinyLlama can be done on platforms like Google Colab Pro for about $10 to $20 per month. In contrast, running the GPT-4 API at scale can cost $100 to $200 or more per month per user, depending on usage.

Operational costs also favor SLMs due to their lower computational demands. Deploying and running LLMs at scale incurs higher infrastructure costs for every processed query. On the other hand, SLMs, with their efficient inference, have lower operational costs, making them a more affordable option for many high-volume, real-time applications. Although LLMs excel in complex, multi-step tasks thanks to their extensive training on varied datasets, SLMs can achieve similar performance in specific applications when properly fine-tuned. Techniques like model distillation, parameter sharing, and quantization have been created to reduce model size without significantly affecting performance. For instance, MobileBERT, a smaller version of BERT, delivers competitive results on benchmarks while being 4.3 times smaller and 5.5 times faster than BERTBASE\cite{sun2020mobilebert}.

SLMs may not perform as well in benchmarks, but they can be fine-tuned for specific repetitive tasks and used as part of a larger system as a group of agents. This allows for using the strengths of SLMs—like efficiency, speed, and scalability—while breaking apart complex tasks into manageable subtasks for individual SLMs. This strategy is also present in reasoning LLMs, where an LLM generates extra tokens to break down a complex task into simpler ones. However, since this process is internal to the LLM, a consistent chain of thought is not guaranteed. In contrast, when creating an AI system with SLMs, we can experiment with and fine-tune our agents to address complex problems in a more repeatable way. This approach leads to more robust, reliable, and efficient solutions.

In summary, SLMs provide a strong alternative to LLMs by offering cost-effective, low-latency, and hardware-efficient solutions, particularly for specialized tasks and real-time applications. Their benefits in deployment flexibility and energy efficiency make them ideal for organizations looking to implement AI capabilities without the large investments typical of LLMs.

\section{Dataset}

In our study, we used a proprietary data set from our sales outreach platform. This data set includes key components of sales campaigns, including value propositions, research goals, identified pain points, and specific sales campaign instructions provided by customers to tailor their outreach efforts. Because these data come from real use cases, they provide a solid foundation for testing how well Large Language Models (LLMs) and Small Language Models (SLMs) perform in actual sales situations. Using unique instructions given by the user is especially important because they have a strong impact on how the model behaves, even when we keep the main system prompt and parameters the same. This highlights how crucial it is to provide context-specific guidance for language models to obtain better results. This finding aligns with other research showing that tailored instructions help models better understand and deliver what users want. \cite{hui2024smaller}

We will be comparing LLMs and SLMs against each other on two tasks:
\subsection{Drafting Sales Outreach}

Large Language Models (LLMs), trained on extensive and diverse datasets, excel in generating coherent and contextually rich text. This capability makes them particularly effective for crafting personalized and engaging sales outreach emails and messages. Their proficiency in understanding nuanced language and producing human-like responses can significantly enhance the effectiveness of sales communications. However, these advantages come with increased computational demands and potential latency issues.

In contrast, Small Language Models (SLMs), when fine-tuned on domain-specific data, can perform comparably to LLMs in generating outreach emails. Their smaller size allows for faster inference and lower resource consumption, making them more practical for real-time applications. Recent research indicates that fine-tuning SLMs for domain-specific tasks can yield quality improvements of approximately 10\% over prompting LLMs, particularly for tasks requiring structured outputs.\cite{ayala2025finetuneslmpromptllm}

\subsubsection*{Evaluating Model Performance}

To evaluate the performance of these models in real-world scenarios, we will measure key email outreach metrics, including:
\begin{itemize}

\item Open Rate: The percentage of recipients who open the email, indicating the effectiveness of the subject lines and the reputation of the sender.
\item Click-Through Rate (CTR): The percentage of recipients who click on links within the email, reflecting the relevance and appeal of the content and calls-to-action.
\item Reply Rate: The percentage of emails that receive a response, which assesses the level of engagement and effectiveness of the message.
\item Unsubscribe Rate: The percentage of recipients who opt out of the email list after receiving the message, providing information on content relevance and audience satisfaction.
\end{itemize}
These metrics offer a realistic performance comparison based on human responses to the sales campaigns, enabling a comprehensive assessment of the effectiveness of the models in practical applications.

\subsection{Conducting Web Research Using Headless Browsers}

Large Language Models (LLMs) possess comprehensive language understanding capabilities, enabling them to interpret and generate complex queries for effective information retrieval. Their extensive training on diverse datasets allows them to navigate and extract relevant information from a wide array of web sources. However, deploying LLMs for such tasks demands substantial computational resources, which may not be feasible for all organizations.

In contrast, Small Language Models (SLMs), when fine-tuned on domain-specific data, can perform web research tasks efficiently. Their reduced computational requirements make them suitable for integration with headless browsers, facilitating automated web navigation and data extraction. This approach enables real-time applications with lower resource consumption. 

\subsubsection*{Evaluating Model Performance and Factual Accuracy}

To assess the performance of these models in web research tasks, we will employ BERTScore, an evaluation metric that leverages pre-trained contextual embeddings from BERT to measure the similarity between generated text and reference text. BERTScore computes precision, recall, and F1 scores by matching words in candidate and reference sentences based on cosine similarity, providing a nuanced evaluation of text generation quality.\cite{zhang2019bertscore}

Additionally, we will validate the factual accuracy of the model-generated responses by cross-referencing the URLs visited during the research process. This involves extracting the text from the visited web pages and comparing it with the model's output to ensure consistency and correctness. By aligning the generated content with the source material, we can identify discrepancies and assess the reliability of the information provided by the models.
\section{Methodology}
The methodology followed a clear process for behavioral mimicking. In the first phase, we carefully chose both teacher models (state-of-the-art LLMs) and learner models (computationally efficient SLMs). After the selection, the teacher LLMs generated a synthetic dataset designed for our target applications. We then underwent a critical human-in-the-loop (HITL) verification stage. Human reviewers assessed each output for quality, relevance, and accuracy. The high-quality data we obtained was used to fine-tune the learner SLMs, resulting in our "Trained Miniatures." The final evaluation was practical. We measured the business impact by A/B testing the Trained Miniatures against the baseline LLMs on key performance indicators like click rate, sign-up rate, reply rate, and final conversion rate.

\subsection{Model Selection}
4 closed-source LLMs from 2 providers were selected as the teachers and 8 open-source SLMs of parameter space scaling from 1B to 8B parameters were chosen.
\subsubsection*{Training Models}
OpenAI:
\begin{itemize}
    \item GPT - 4o\cite{openai2024gpt4ocard}
    \item GPT - 4.1\cite{openai_gpt41_2024}
\end{itemize}
\hspace{0.8em} Anthropic:
\begin{itemize}
    \item Claude - 3.7 Sonnet\cite{anthropic2025claude3p7sonnet}
    \item Claude - 4.0 Sonnet\cite{anthropic2025claudesonnet4}
\end{itemize}

\subsubsection*{Learning Models}
Google --- Gemma:
\begin{itemize}
    \item Gemma-3-12B-it (12B parameters, 32K context)
    \item Gemma-3-4B-it (4B parameters, 16K context)
    \item Gemma-3-1B-it (1B parameters, 16K context)
\end{itemize}
\hspace{1.1em}Alibaba --- Qwen
\begin{itemize}
    \item Qwen/Qwen3-4B (4B parameters, 8K context)
    \item Qwen/Qwen3-1.7B (1.7B parameters, 8K context)
    \item Qwen2-1.5B-Instruct (1.5B parameters, 8K context)
\end{itemize}
\hspace{0.8em}Meta --- Llama
\begin{itemize}
    \item Llama 3.2 Instruct (3B) (3B, 8k context)
    \item Llama 3.2 Instruct (1B) (1B, 8k context)
\end{itemize}

\hspace{-1.1em}This selection allowed us to cover a significant number of models, with varying architectures and parameter spaces ranging from 1B to 12B and context length ranging from 8k tokens to 32 tokens.

\subsection{Fine-tuning}
To investigate the abilities and limits of Small Language Models (SLMs), we used two different adaptation methods for each model. The first method was a complete fine-tuning process, where we updated all model parameters. The second method used Parameter-Efficient Fine-Tuning (PEFT), specifically Low-Rank Adaptation (LoRA), which changed only a small part of the model's parameters. This two-pronged approach helped us compare model performance with different amounts of computing power and memory. It also clarified how these tuning methods affected the model's ability to generalize tasks effectively.
\subsubsection{LoRA Finetuning approach}
We used Low-Rank Adaptation (LoRA) (Hu et al., 2021) for parameter-efficient finetuning of all the SLMs (Small Language models).  LoRA decomposes weight updates into low-rank matrices, reducing trainable parameters by 99\% while maintaining model performance. This is also called as parameter-efficient finetuning.

\begin{figure}[htbp]
\centerline{\includegraphics[scale=0.35]{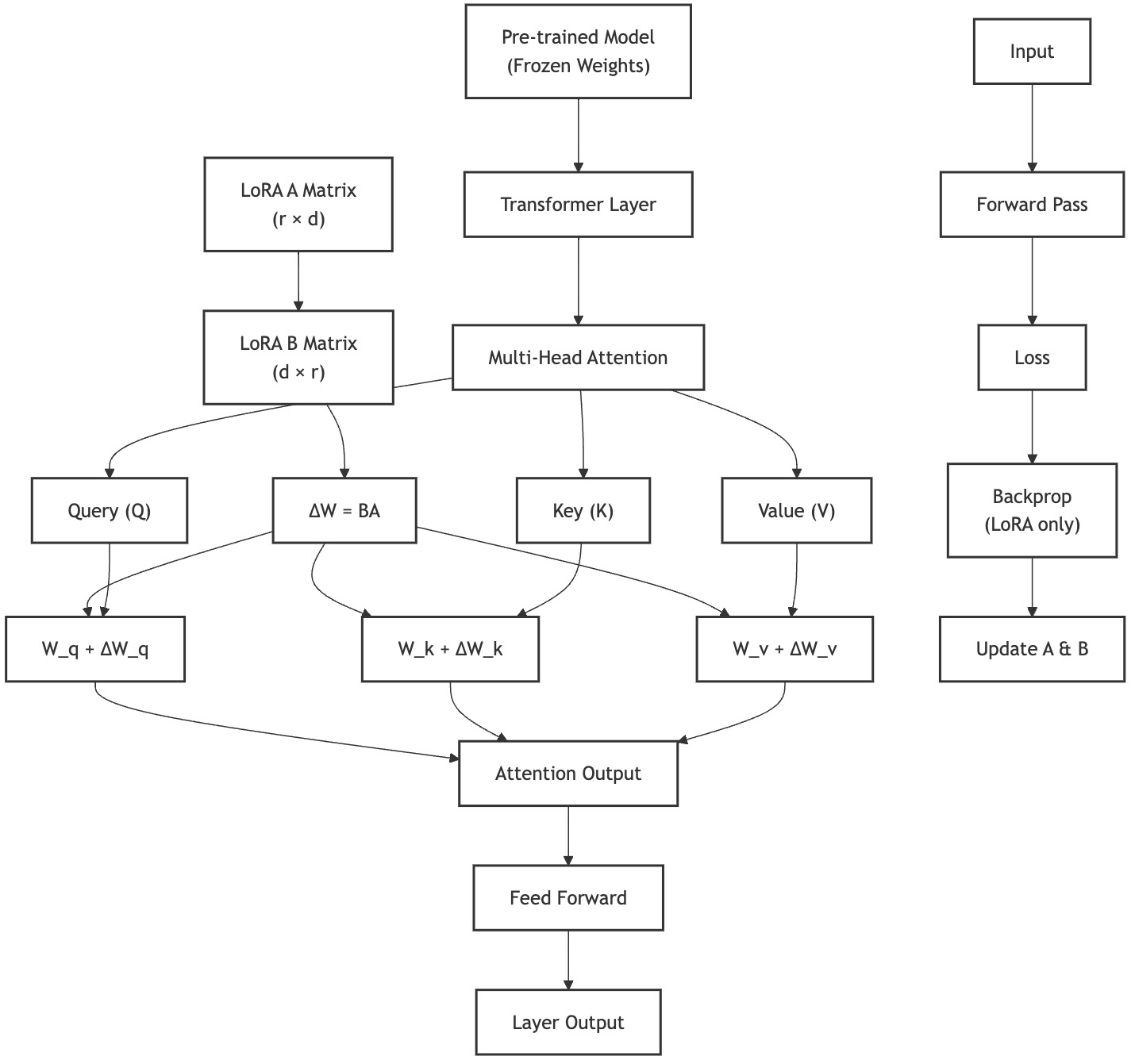}}
\caption{LoRA Schematic View.}
\label{Lora}
\end{figure}
\begin{equation}
W = W_0 + BA
\end{equation}
Where $W_0$ represents frozen pre-trained weights, and $B \in \mathbb{R}^{d \times r}$ and $A \in \mathbb{R}^{r \times k}$ are trainable low-rank matrices with rank $r \ll \min(d,k)$.
\subsubsection*{LoRA Configuration:}
$\text{Rank}(r) = 16$ for models $\le 3\text{B}$ parameters, $32$ for models $> 3\text{B}$ parameters, and $\alpha = 32$ (scaling factor), and $\text{dropout} = 0.05$.
\subsubsection*{Training type:}
Automatic Mixed Precision (AMP) -  (Micikevicius et al., 2017) AMP dynamically uses 16-bit (half-precision) floating-point arithmetic for forward and backward passes while maintaining 32-bit precision for loss scaling and parameter updates. This approach reduces GPU memory consumption by approximately $50\%$ and accelerates training by $1.5-2\times$ on modern hardware, enabling us to train larger batch sizes and longer sequences within memory constraints. The automatic loss scaling mechanism prevents gradient underflow that can occur with reduced precision, ensuring training stability while maintaining model convergence quality equivalent to full 32-bit precision training.
\subsubsection*{Training Configuration}

All the models were finetuned using the below hyperparameters, optimized through preliminary experiments:
\begin{itemize}
    \item  Optimizer: AdamW with $\beta_1=0.9, \beta_2=0.999$
    \item Learning Rate: $2^{e-4}$ with cosine annealing schedule
    \item Batch Size: Adaptive based on model size ($8-32$ per GPU)
    \item Gradient Accumulation: 4 steps
    \item Max Sequence Length: $2048$ tokens
    \item Training Epochs: $3-5$ (determined by validation loss convergence)
    \item Warmup Steps: $10\%$ of total training steps
\end{itemize}

\subsubsection{Full Finetuning}
Full finetuning is a machine learning technique in which all the parameters of a pre-trained model are updated during training. Unlike parameter-efficient methods (LoRA), every weight in the neural network from input embeddings to output layers becomes trainable and gets modified based on the new training data i.e. our sales/marketing dataset here. The process takes a pre-trained model and continues training with all parameters unlocked. Gradients flow backward through every layer during backpropagation, updating billions of parameters simultaneously. The model essentially rewrites its internal representations to better fit the target task, starting from the pre-trained weights as initialization.

Every model evaluated in this study was fine-tuned twice, once using full fine-tuning and once using LoRA, on the same training data and hyperparameter schedules (where applicable). This ensured a controlled comparison of task performance, convergence behavior, and inference efficiency between both training paradigms.\subsubsection*{Full Fine-tuning Configuration:}
We conducted full fine-tuning experiments using significantly different hyperparameter settings compared to our LoRA approach.
\subsubsection*{Memory Optimization Techniques}
Due to the memory constraints from updating all parameters in full fine-tuning, we employed several advanced techniques:
\begin{itemize}
    \item \textbf{Automatic Mixed Precision (AMP)}: Utilizing FP16/BF16 (Micikevicius et al., 2017) to reduce memory consumption and accelerate training.
    \item \textbf{Gradient Checkpointing}: Recomputing activations during the backward pass to reduce memory footprint (Chen et al., 2016).
    \item \textbf{ZeRO Sharding}: Employed for our larger 14\text{B} parameter models to distribute optimizer states, gradients, and parameters across devices (Rajbhandari et al., 2020).
\end{itemize}
We observed memory requirements scaling dramatically for full fine-tuning, needing 80-160GB for 1\text{B}-14\text{B} models in our experimental setup.

\subsubsection*{Full Fine-tuning Hyperparameters}
We optimized the full fine-tuning process using the following hyperparameters, which were significantly different from our LoRA approach to prevent catastrophic forgetting and manage resources:
\begin{itemize}
    \item Optimizer: AdamW (Loshchilov \& Hutter, 2017) with standard $\beta_1=0.9, \beta_2=0.999$
    \item Learning Rate: $1 \times 10^{-5}$ to $5 \times 10^{-5}$ (10-100x lower than our LoRA configuration)
    \item Batch Size per GPU: $1-4$
    \item Gradient Accumulation Steps: $8-32$ (achieving effective batch sizes of $32-128$)
    \item Training Epochs: $1-3$ (fewer than our LoRA experiments)
    \item Weight Decay: $0.01-0.1$
    \item Dropout: $0.1$
    \item Gradient Clipping: $1.0$
    \item Warmup Steps: $5-10\%$ of total training steps for stable training
\end{itemize}
The overall full fine-tuning process required 10-50x more compute time and resources compared to our LoRA fine-tuning approach.

\begin{figure}[h!]
\centerline{\includegraphics[scale=0.35]{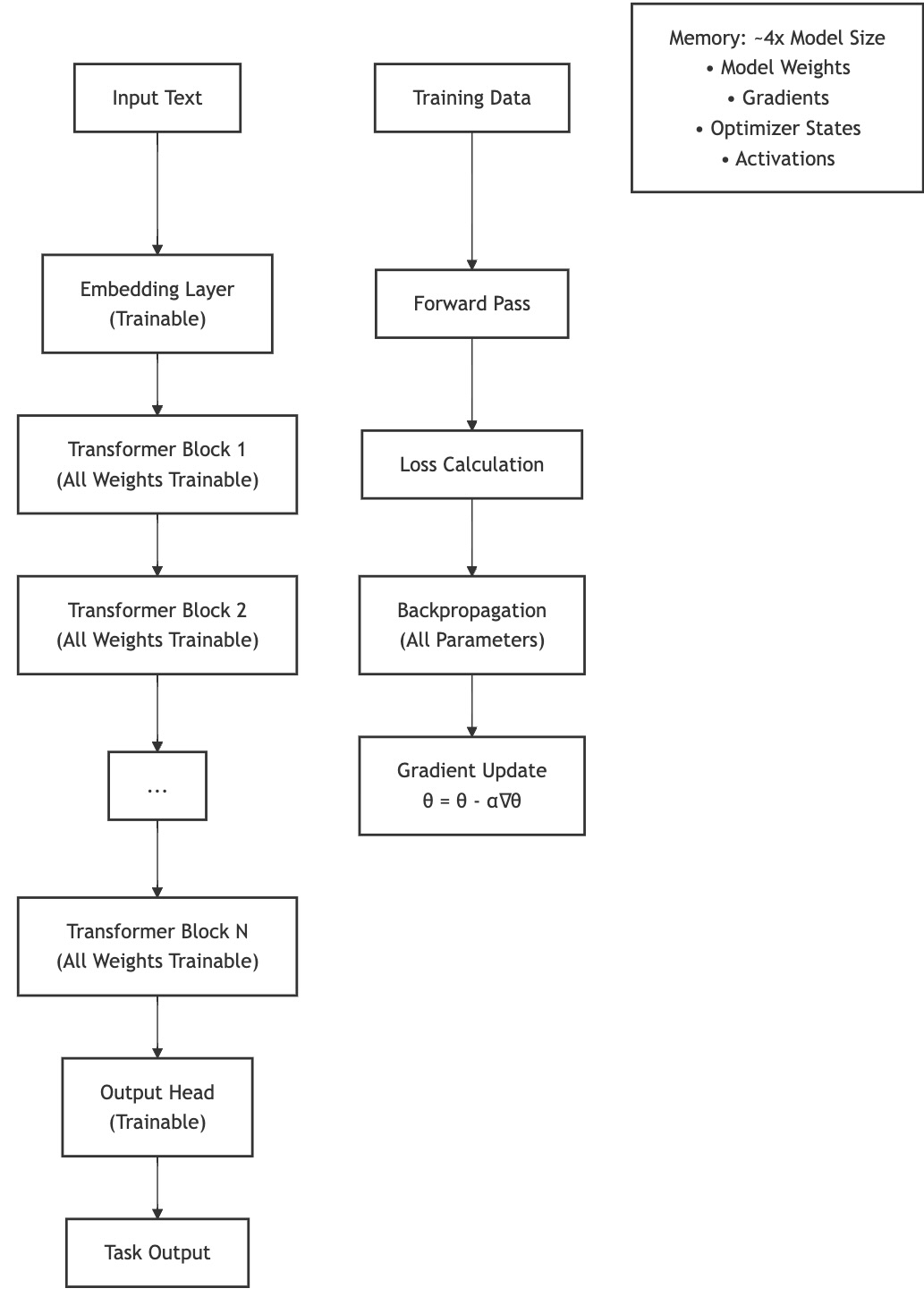}}
\caption{Full Finetuning Schematic View.}
\label{fft}
\end{figure}
\pagebreak
\section{Agentic Cold-Outreach}
The cold-outreach campaign is an agent controlled campaign where emails and LinkedIn Messages are sent out in a preconfigured sequence. The number of steps, the delay between each step and, outreach instructions and outreach type(LinkedIn/Email) can be configured. \textit{Fig. 3.} shows a sample configuration.

\begin{figure}[h!]
\centerline{\includegraphics[scale=0.14]{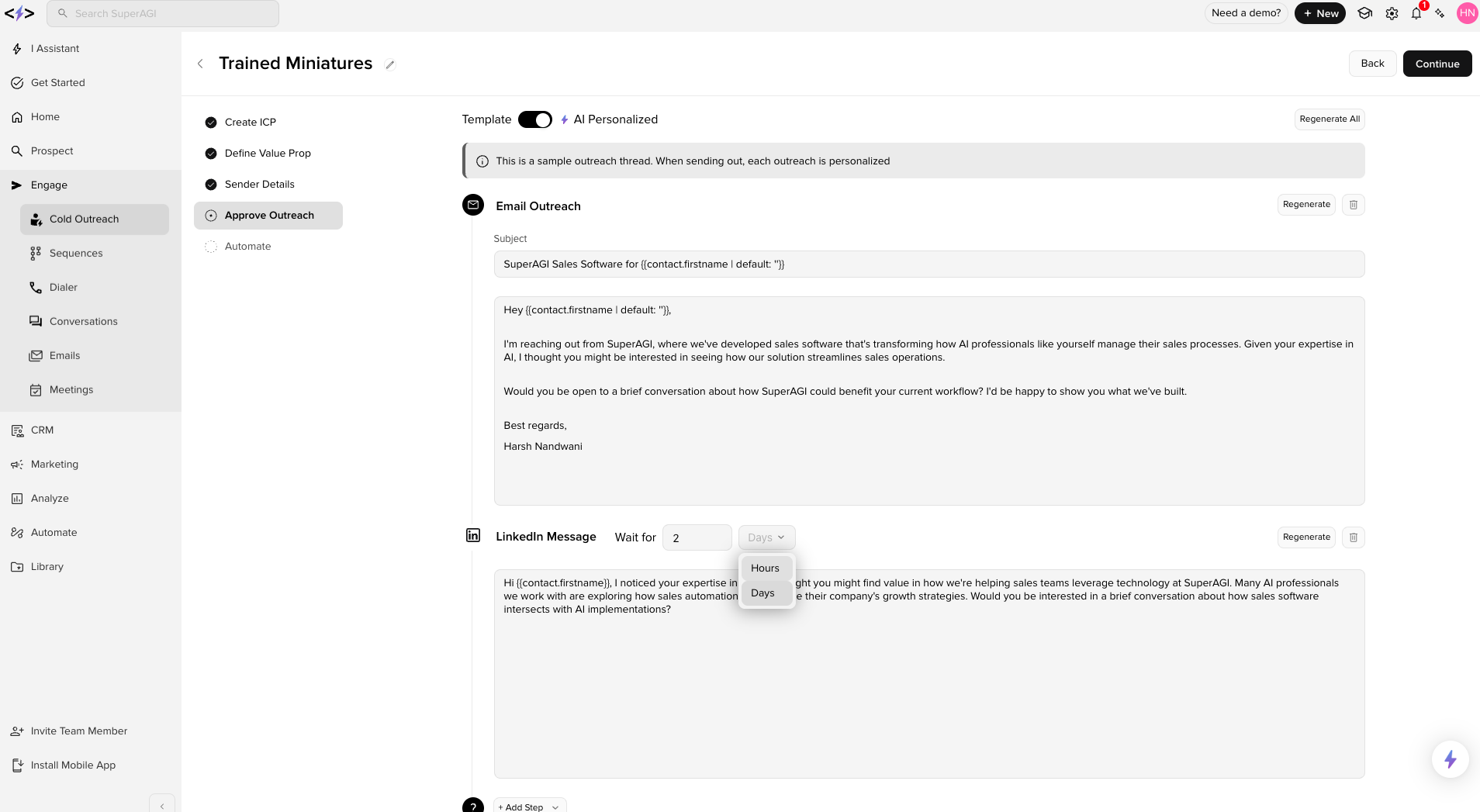}}
\caption{Sample Cold-Outreach Configuration.}
\label{supersales}
\end{figure}

These instructions are stored in the agent's memory. At the time of creation, an initial draft is generated on the basis of these configured values. When a new lead is added to the campaign, market research is performed per lead and added to the agent's memory. The lead associations stored also include all previous communications sent, as well as, responses received to craft the most suitable replies and follow-ups.

\begin{figure}[h!]
\centerline{\includegraphics[scale=0.12]{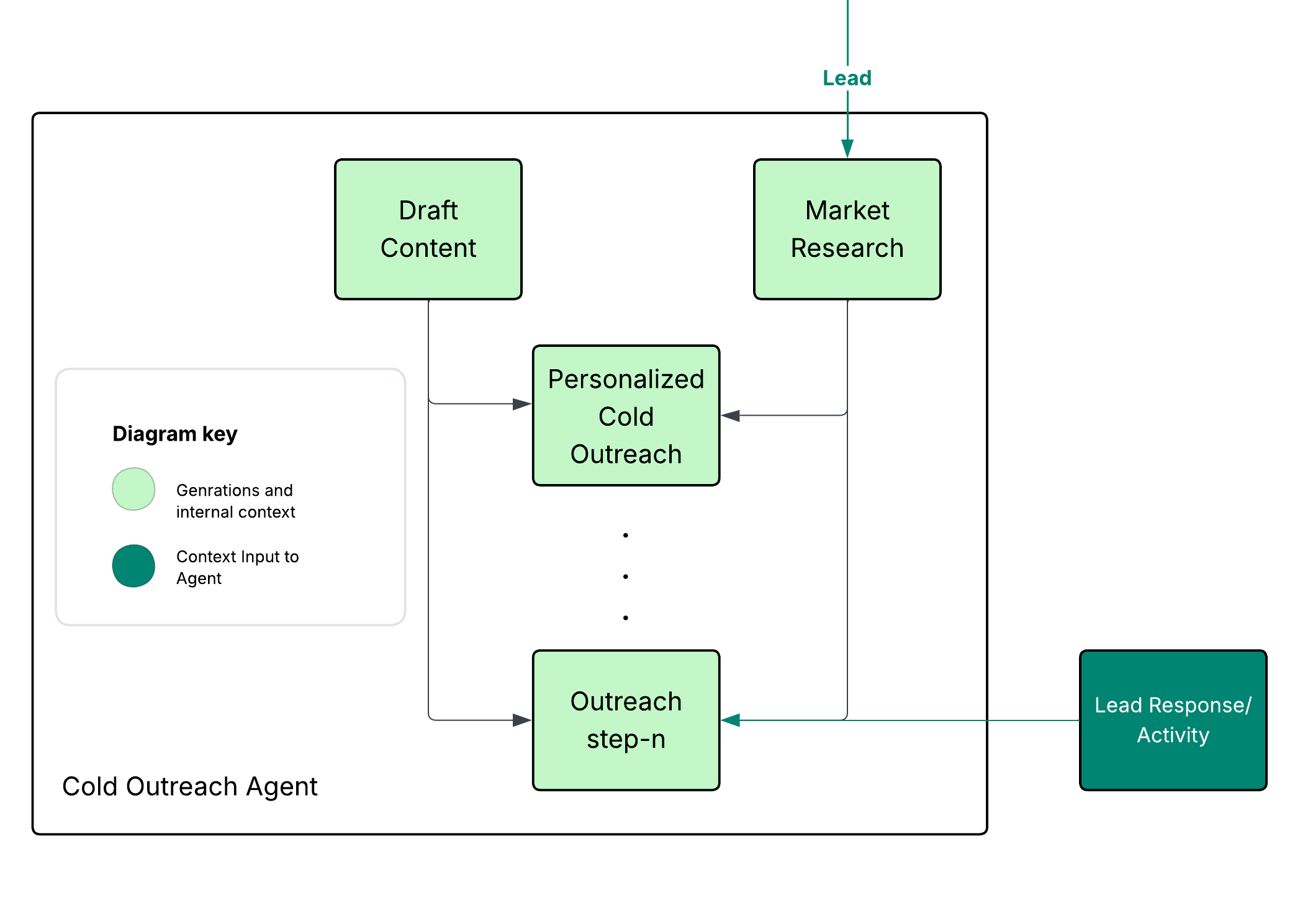}}
\caption{Cold-Outreach Agent Structure.}
\label{agent}
\end{figure}

\section{Evaluation}

\subsection{Overall Observations}
\textit{Table 2} presents a comprehensive comparison of baseline LLMs, LoRA finetuned, and fully finetuned SLM models across email effectiveness metrics: CTR (Click-Through Rate), Open Rate, and Response Rate. We observe many different trends when analyzing these results:

\begin{table*}[h!]
\small
\label{email_metrics}
\centering
\begin{tabularx}{\textwidth}{l l l l c c c}
\toprule
\textbf{Model Name} & \textbf{Model Type} & \textbf{Parameters} & \textbf{Context Length} & \textbf{CTR (\%)} & \textbf{Open Rate (\%)} & \textbf{Response Rate (\%)} \\
\midrule
\multicolumn{7}{l}{\textbf{Baseline Models}} \\
GPT-4o & Baseline LLM & $>$100B & 128K & 3.2 & 33.2 & 5.7 \\
GPT-4.1 & Baseline LLM & $>$100B & 128K & 3.4 & 34.1 & 6.2 \\
Claude-3.7-Sonnet & Baseline LLM & $>$100B & 200K & 3.1 & 32.8 & 5.4 \\
Claude-4-Sonnet & Baseline LLM & $>$100B & 200K & 3.5 & 35.2 & 6.5 \\
\addlinespace
\multicolumn{7}{l}{\textbf{LoRA Finetuned Models}} \\
Gemma-3-12B-it (LoRA) & LoRA Finetuned & 12B & 32K & 3.3 & 28.8 & 5.9 \\
Gemma-3-4B-it (LoRA) & LoRA Finetuned & 4B & 16K & 3.0 & 27.5 & 5.1 \\
Gemma-3-1B-it (LoRA) & LoRA Finetuned & 1B & 16K & 2.7 & 26.8 & 4.2 \\
Qwen3-4B (LoRA) & LoRA Finetuned & 4B & 8K & 2.9 & 27.2 & 4.9 \\
Qwen3-1.7B (LoRA) & LoRA Finetuned & 1.7B & 8K & 2.8 & 26.6 & 4.5 \\
Qwen2-1.5B-Instruct (LoRA) & LoRA Finetuned & 1.5B & 8K & 2.6 & 26.1 & 4.3 \\
Llama 3.2 Instruct 3B (LoRA) & LoRA Finetuned & 3B & 8K & 3.1 & 28.9 & 5.2 \\
Llama 3.2 Instruct 1B (LoRA) & LoRA Finetuned & 1B & 8K & 2.5 & 26.4 & 3.9 \\
\addlinespace
\multicolumn{7}{l}{\textbf{Full Finetuned Models}} \\
Gemma-3-12B-it (Full) & Full Finetuned & 12B & 32K & 3.4 & 31.2 & 6.1 \\
Gemma-3-4B-it (Full) & Full Finetuned & 4B & 16K & 3.2 & 30.6 & 5.6 \\
Gemma-3-1B-it (Full) & Full Finetuned & 1B & 16K & 2.9 & 29.0 & 4.8 \\
Qwen3-4B (Full) & Full Finetuned & 4B & 8K & 3.1 & 30.3 & 5.4 \\
Qwen3-1.7B (Full) & Full Finetuned & 1.7B & 8K & 3.0 & 29.7 & 5.1 \\
Qwen2-1.5B-Instruct (Full) & Full Finetuned & 1.5B & 8K & 2.8 & 29.3 & 4.9 \\
Llama 3.2 Instruct 3B (Full) & Full Finetuned & 3B & 8K & 3.2 & 30.4 & 5.5 \\
Llama 3.2 Instruct 1B (Full) & Full Finetuned & 1B & 8K & 2.7 & 29.5 & 4.4 \\
\bottomrule
\hspace{3.5pt}
\end{tabularx}
\caption{Comparison of model performance on email marketing metrics.}
\end{table*}

\begin{itemize}
    \item \textbf{Baseline LLMs}: As expected, large-scale proprietary models such as GPT-4o, GPT-4.1, and Claude-4 exhibit the strongest performance across all metrics, setting an upper limit on what is currently achievable for email tasks, with response rates exceeding 6\% and open rates above 33\%.
    \item \textbf{Full Finetuned 12B+ Models}: Models such as Gemma-3-12B-it (full) closely match baselines, achieving a CTR of 3.4\%, an open rate of 31.2\%, and a response rate of 6.1\%. This highlights the effectiveness of task-specific fine-tuning for large open-source models.
    \item \textbf{LoRA Finetuned 12B+ Models}: These models, for example, Gemma-3-12B-it (LoRA), also perform competitively (CTR 3.3\%, response rate 5.9\%) - demonstrating that parameter-efficient fine-tuning methods can capture much of the benefit of full fine-tuning but at significantly lower computational cost.
    \item \textbf{4B Parameter Range}: Both LoRA and fully fine-tuned models in this class (e.g., Qwen3-4B, Gemma-3-4B) show a moderate reduction in effectiveness, with response rates between 5.1 and 5. 6\%. However, the performance drop is not proportional to the significant reduction in parameter count, indicating a good trade-off for resource-constrained deployments.
    \item \textbf{1–3B Models}: Models below the 3B parameters, while considerably more cost-effective, tend to exhibit a sharper decline across all metrics, particularly in response rate and open rate. This suggests inherent limitations in capturing the subtleties required for engaging, personalized email content.
\end{itemize}

\begin{table*}[t]
\centering
\footnotesize
\begin{tabularx}{\textwidth}{l l c c c c c c c}
\toprule
\textbf{Model} & \textbf{Type} & \textbf{Params} & \textbf{Ctx} & \textbf{BERT F1} & \textbf{ROUGE-L} & \textbf{Fact. (\%)} & \textbf{Human Rel. (\%)} & \textbf{Human Comp. (\%)} \\
\midrule
\multicolumn{9}{l}{\textbf{Baseline}} \\
GPT-4o & Base LLM & $>$100B & 128K & 0.892 & 0.447 & 89 & 92 & 89 \\
GPT-4.1 & Base LLM & $>$100B & 128K & 0.901 & 0.453 & 91 & 93 & 90 \\
Claude-3.7-Sonnet & Base LLM & $>$100B & 200K & 0.887 & 0.441 & 89 & 90 & 88 \\
Claude-4-Sonnet & Base LLM & $>$100B & 200K & 0.905 & 0.459 & 92 & 94 & 92 \\
\addlinespace
\multicolumn{9}{l}{\textbf{Full Finetuned}} \\
Gemma-3-12B Full & Full FT & 12B & 32K & 0.871 & 0.426 & 86 & 88 & 85 \\
Gemma-3-4B Full & Full FT & 4B & 16K & 0.854 & 0.407 & 83 & 85 & 82 \\
Gemma-3-1B Full & Full FT & 1B & 16K & 0.829 & 0.379 & 79 & 81 & 78 \\
Qwen3-4B Full & Full FT & 4B & 8K & 0.850 & 0.402 & 82 & 84 & 81 \\
Qwen3-1.7B Full & Full FT & 1.7B & 8K & 0.836 & 0.385 & 80 & 82 & 79 \\
Qwen2-1.5B Instruction Full & Full FT & 1.5B & 8K & 0.823 & 0.371 & 78 & 80 & 77 \\
Llama 3.2-3B Full & Full FT & 3B & 8K & 0.841 & 0.394 & 81 & 83 & 80 \\
Llama 3.2-1B Full & Full FT & 1B & 8K & 0.817 & 0.358 & 76 & 78 & 75 \\
\addlinespace
\multicolumn{9}{l}{\textbf{LoRA Finetuned}} \\
Gemma-3-12B LoRA & LoRA FT & 12B & 32K & 0.862 & 0.413 & 83 & 85 & 82 \\
Gemma-3-4B LoRA & LoRA FT & 4B & 16K & 0.839 & 0.388 & 79 & 82 & 78 \\
Gemma-3-1B LoRA & LoRA FT & 1B & 16K & 0.811 & 0.357 & 75 & 77 & 74 \\
Qwen3-4B LoRA & LoRA FT & 4B & 8K & 0.835 & 0.381 & 78 & 80 & 77 \\
Qwen3-1.7B LoRA & LoRA FT & 1.7B & 8K & 0.819 & 0.364 & 76 & 78 & 75 \\
Qwen2-1.5B-Inst LoRA & LoRA FT & 1.5B & 8K & 0.807 & 0.352 & 74 & 76 & 73 \\
Llama 3.2-3B LoRA & LoRA FT & 3B & 8K & 0.833 & 0.385 & 78 & 81 & 77 \\
Llama 3.2-1B LoRA & LoRA FT & 1B & 8K & 0.800 & 0.341 & 72 & 75 & 71 \\

\bottomrule
\hspace{3.5pt}
\end{tabularx}
\caption{Market research model metrics. 
FT = Finetuned; LoRA = Low-Rank Adaptation; Params = parameters; Ctx = context length; Fact. = Factual Accuracy; Human Rel./Comp. = Human Relevance/Completeness}
\end{table*}

\subsection{Cost-Efficiency and Practical Implications}
A key motivation for exploring LoRA and lower-parameter models is to reduce inference cost and deployment latency. Our findings suggest:

\begin{itemize}
    \item \textbf{LoRA Finetuning}: Offers substantial cost savings by only updating a fraction of the parameters while retaining most of the performance, especially for medium-sized models (4B–12B). This is well suited for organizations that require frequent retraining or on-demand inference.
    \item \textbf{Full Finetuning}: Delivers higher performance at the expense of increased compute, particularly beneficial when maximizing engagement is critical.
    \item \textbf{Model Size Trade-off}: For applications prioritizing cost, 3–4B models offer a balanced compromise, whereas sub-3B models may be best reserved for low-priority or high-volume use cases.
\end{itemize}

\subsection{Notable Insights}
\begin{itemize}
    \item \textbf{LoRA vs. Full Finetuning}: The performance gap between LoRA and full finetuned models is typically less than 0.2–0.3\% across metrics in the same parameter class, advocating for LoRA in resource-sensitive contexts.
    \item \textbf{Context Length}: While all top-performing models support longer context windows ($\geq$ 16K), there is no substantial drop-off in metrics for 8K models within the same parameter range, suggesting diminishing returns beyond a certain threshold for this task.
    \item \textbf{Baseline Ceiling}: Proprietary LLMs still retain an advantage, especially in generating highly engaging and contextually tailored content, but the gap is narrowing with recent advances in finetuning large-scale open-source models.
\end{itemize}

\section{Metrics Explanation}

\subsection{Automated ML Metrics}
\begin{itemize}
    \item \textbf{BERTScore (F1)}: Semantic similarity between generated and reference market research summaries using BERT embeddings (range: 0--1).
    \item \textbf{ROUGE-L}: Longest common subsequence-based metric for summary quality (range: 0--1).
    \item \textbf{Factual Accuracy (\%)}: Score from automated fact-checking models comparing output with source information.
\end{itemize}

\subsection{Human Evaluation Metrics}
\begin{itemize}
    \item \textbf{Human Relevance (\%)}: Human rating of how relevant model insights are to research objectives.
    \item \textbf{Human Completeness (\%)}: Human assessment of the comprehensiveness of the market research summary.
\end{itemize}

\section{Metric Calculation Methodology}

\subsection{Automated ML Metrics}
\subsubsection*{BERTScore (F1)}
\textbf{Calculation Process:}
\begin{enumerate}
    \item Encode generated and reference summaries using a pretrained BERT model.
    \item Compute token-level cosine similarities between embeddings.
    \item Calculate precision: \( P = (1/|x|) \sum \) maximum similarity for each token in generated text.
    \item Calculate recall: \( R = (1/|y|) \sum \) maximum similarity for each token in reference text.
    \item \( F1 = 2 \cdot (P \cdot R) / (P + R) \)
\end{enumerate}

\begin{table*}[h!]
\small
\centering
\begin{tabularx}{\textwidth}{ l l c c c c }
\toprule
\textbf{Model Name} & \textbf{Model Type} & \textbf{Total System Cost} & \textbf{CTR (\%)} & \textbf{Open Rate (\%)} & \textbf{Response Rate (\%)} \\
\midrule
\multicolumn{6}{l}{\textbf{Baseline Models}} \\
\addlinespace
GPT-4o                 & Agentic Baseline & \textbf{\$0.1383} & 3.7 & 36.5 & 6.4 \\
GPT-4.1                & Agentic Baseline & \textbf{\$0.1106} & 3.9 & 37.4 & 6.9 \\
Claude-3.7-Sonnet      & Agentic Baseline & \textbf{\$0.2007} & 3.6 & 36.1 & 6.1 \\
Claude-4-Sonnet        & Agentic Baseline & \textbf{\$0.2007} & 4.0 & 38.7 & 7.2 \\
\addlinespace
\multicolumn{6}{l}{\textbf{LoRA Finetuned Models}} \\
\addlinespace
Gemma-3-12B-it (LoRA)        & Agentic LoRA & \textbf{\$0.0071} & 3.5 & 31.0 & 6.0 \\
Gemma-3-4B-it (LoRA)         & Agentic LoRA & \textbf{\$0.0035} & 3.3 & 30.0 & 5.6 \\
Gemma-3-1B-it (LoRA)         & Agentic LoRA & \textbf{\$0.0012} & 3.0 & 29.0 & 4.8 \\
Qwen3-4B (LoRA)              & Agentic LoRA & \textbf{\$0.0035} & 3.2 & 29.5 & 5.4 \\
Qwen3-1.7B (LoRA)            & Agentic LoRA & \textbf{\$0.0018} & 3.1 & 29.0 & 5.1 \\
Qwen2-1.5B-Instruct (LoRA)   & Agentic LoRA & \textbf{\$0.0018} & 3.0 & 28.7 & 4.9 \\
Llama 3.2 Instruct 3B (LoRA) & Agentic LoRA & \textbf{\$0.0029} & 3.3 & 30.5 & 5.7 \\
Llama 3.2 Instruct 1B (LoRA) & Agentic LoRA & \textbf{\$0.0012} & 2.9 & 24.3 & 2.5 \\
\addlinespace
\multicolumn{6}{l}{\textbf{Full Finetuned Models}} \\
\addlinespace
Gemma-3-12B-it (Full)        & Agentic Full & \textbf{\$0.0071} & 3.7 & 34.0 & 6.5 \\
Gemma-3-4B-it (Full)         & Agentic Full & \textbf{\$0.0035} & 3.6 & 33.2 & 6.1 \\
Gemma-3-1B-it (Full)         & Agentic Full & \textbf{\$0.0012} & 3.3 & 31.5 & 5.3 \\
Qwen3-4B (Full)              & Agentic Full & \textbf{\$0.0035} & 3.5 & 32.8 & 5.9 \\
Qwen3-1.7B (Full)            & Agentic Full & \textbf{\$0.0018} & 3.4 & 32.2 & 5.6 \\
Qwen2-1.5B-Instruct (Full)   & Agentic Full & \textbf{\$0.0018} & 3.2 & 31.7 & 5.4 \\
Llama 3.2 Instruct 3B (Full) & Agentic Full & \textbf{\$0.0029} & 3.5 & 32.9 & 6.0 \\
Llama 3.2 Instruct 1B (Full) & Agentic Full & \textbf{\$0.0012} & 3.1 & 26.0 & 3.2 \\

\bottomrule
\hspace{3.5pt}
\end{tabularx}

\caption{Cold-Outreach Agent metrics, per lead (average for 2000 leads across 10 campaigns with mean 4 steps)}
\end{table*}

\textbf{Implementation:} Uses \texttt{bert-base-uncased}, applies IDF weighting, and rescales with baseline scores for interpretability.

\textbf{Validation:}
\begin{itemize}
    \item Correlation with human relevance: \( r = 0.73 \) (p~$<$~0.001, n=500)
    \item Validated on 100 held-out test summaries, across three domains (tech, healthcare, finance)
    \item Inter-domain consistency: \( r = 0.68{-}0.78 \)
\end{itemize}

\subsubsection*{ROUGE-L}
\textbf{Calculation Process:}
\begin{enumerate}
    \item Find the longest common subsequence (LCS) between generated and reference summaries.
    \item Precision: \( P = \text{LCS\_length} / \text{generated\_length} \)
    \item Recall: \( R = \text{LCS\_length} / \text{reference\_length} \)
    \item F-measure: \( \text{ROUGE-L} = \frac{(1+\beta^2)RP}{R+\beta^2P} \), where \( \beta = 1.2 \)
\end{enumerate}

\textbf{Features:}
\begin{itemize}
    \item Captures sentence-level structure and word order.
    \item More suitable for summary coherence than ROUGE-1/2.
\end{itemize}

\subsubsection*{Factual Accuracy (\%)}
\textbf{Calculation Process:}
\begin{enumerate}
    \item Use NER to extract factual claims from model output.
    \item Cross-reference claims against source documents via automated fact-checking.
    \item Classify as Supported/Contradicted/Unverifiable.
    \item Score: \( \frac{\text{Supported Claims}}{\text{Total Verifiable Claims}} \times 100 \)
\end{enumerate}

\textbf{Implementation:} spaCy NER for entity/dates/numerical facts; fact verification via exact match \& semantic similarity; manual validation on 100 samples (\(89\%\) agreement), focused on verifiable claims.

\textbf{Validation:}
\begin{itemize}
    \item Human expert validation on 100 randomly selected samples.
    \item Inter-annotator agreement: \( \kappa = 0.82 \).
    \item Subjective claims/opinions excluded from accuracy score.
\end{itemize}

\subsection{Human Evaluation Metrics}

\subsubsection*{Human Relevance (\%)}
\textbf{Evaluation Process:}
\begin{enumerate}
    \item Evaluators receive: objectives, source docs, generated summary.
    \item 5-point rating scale:
    \begin{itemize}
        \item 5: Extremely relevant (perfectly addresses objectives)
        \item 4: Highly relevant (minor gaps)
        \item 3: Moderately relevant (core objectives only)
        \item 2: Somewhat relevant (significant gaps)
        \item 1: Not relevant
    \end{itemize}
    \item Convert to percent: \( \text{Score} = \frac{\text{Rating}-1}{4} \times 100 \)
    \item Mean of 3 evaluators per sample (n=500). Inter-annotator agreement: Krippendorff's \( \alpha = 0.74 \).
\end{enumerate}

\subsubsection*{Human Completeness (\%)}
\textbf{Evaluation Process:}
\begin{enumerate}
    \item Checklist covers: executive summary, company background, market analysis, competitors, finance, strategy, supporting data.
    \item Binary (present/absent, with quality threshold) per component; weighted by importance.
    \item Score: \( \frac{\text{Completed Weighted Components}}{\text{Total Possible Score}} \times 100 \)
    \item 3 evaluators/sample; 1,500 assessments; regular calibration and adjudication.
\end{enumerate}

\subsection{Quality Assurance}
\subsubsection*{Inter-Rater Reliability}
\begin{itemize}
    \item Exactly 3 independent evaluators per human metric.
    \item Krippendorff's alpha: \( \alpha = 0.74 \) (threshold $>$ 0.7).
    \item Regular calibration/exemplar review; protocols for high-variance scores.
\end{itemize}
\subsubsection*{Automated Metric Validation}
\begin{itemize}
    \item BERTScore--human relevance: \( r = 0.73 \) (n=500)
    \item ROUGE-L--human completeness: \( r = 0.61 \) (\( \beta=1.2 \))
    \item Factual accuracy--human agreement: \( \kappa = 0.82 \)
    \item Strong performance over multiple domains; best for factual content.
\end{itemize}
\subsubsection*{Evaluation Dataset}
\begin{itemize}
    \item 500+ research samples, annotated by experts, broad industry coverage, regular updates/QA.
\end{itemize}

\section{Key Observations}

\subsection{Performance Tiers}
\begin{enumerate}
    \item \textbf{Baseline LLMs}: Highest scores, especially Claude-4-Sonnet (BERTScore: 0.905, Factual Accuracy: 92\%).
    \item \textbf{Full Finetuned 12B+}: $\sim$96\% of baseline, cost-efficient.
    \item \textbf{Full Finetuned 3--4B}: $\sim$88--94\% of baseline, best cost/perf. balance.
    \item \textbf{LoRA Finetuned 12B+}: $\sim$91--95\% of baseline, $\sim$10$\times$ lower training cost.
    \item \textbf{LoRA 3B}: $\sim$85--92\% of baseline.
    \item \textbf{LoRA Sub-3B}: $\sim$80--89\% of baseline, good for basic research.
\end{enumerate}

\subsection{Cost-Performance Analysis}
\begin{itemize}
    \item \textbf{Training cost}: LoRA $\sim$10$\times$ cheaper than full finetuning (GPU-hours).
    \item \textbf{Inference cost}: SLMs $\sim$8--10$\times$ cheaper than baseline LLMs (per 1M tokens).
    \item \textbf{Training speed}: LoRA $\sim$3--5$\times$ faster.
    \item \textbf{Data}: Full finetuning needs 3--5$\times$ more data.
\end{itemize}

\subsection{Metric Correlations}
\begin{itemize}
    \item \textbf{BERTScore}: Moderate correlation with human relevance ($r = 0.73$).
    \item \textbf{ROUGE-L}: Weaker for human completeness ($r = 0.61$, $\beta = 1.2$ optimised).
    \item \textbf{Factual Accuracy}: Strong alignment with human scores ($\kappa = 0.82$).
    \item \textbf{Human completeness}: Captures coverage better than automated scores.
    \item Metrics best for factual, less so for creative/strategic content.
\end{itemize}

\section{Future Scope}

In this paper, we present our approach to developing a multi-SLM agent framework designed to meet application-specific requirements. Our implementation achieves strong performance while significantly reducing both hardware and cloud costs, making it a highly viable solution for large-scale deployment of AI-powered agent frameworks. Building on the recent advent of open-source models such as Kimi K2 and others, we can enable improved distillation techniques for SLMs. Given the exponential progress in LLM development in recent years, we anticipate that advances in LLM models, particularly open-weight models, will naturally transfer to SLMs. This transfer will make SLMs increasingly competitive and practical, while maintaining their ability to run efficiently on lower-end hardware. One of the rather interesting approaches to distillation for trained miniatures can involves, isolating experts in Mixture-of-Experts(MOE) models and distill the miniatures on experts' weights. These techniques can also help to implemet specialized agents for marketing and customer support use cases. Due to the low cost of these miniatures, they can also sit on top of retrival mechanisms for RAG applications, vetting the retrieved sources; adding another layer of verification to establish reliability at scale. 

\section{Conclusion}
In summary, our evaluation demonstrates that with careful selection and finetuning, open-source foundation models—particularly in the 4B–12B parameter range—can approach the performance of leading proprietary LLMs on email engagement tasks. LoRA finetuning emerges as a compelling strategy for organizations seeking to balance performance and cost, while full finetuning retains its place when maximizing output quality is paramount.

\bibliographystyle{unsrtnat}

\bibliography{TrainedMiniatures}
\end{document}